\documentclass[5p,times]{elsarticle}

\usepackage{graphicx}
\usepackage{caption}
\usepackage{subcaption}
\usepackage{tikz}
\usepackage{soul}
\usepackage{amsmath}
\usepackage{amssymb}
\usepackage{booktabs}

\DeclareMathOperator*{\argmax}{arg\,max}

\usepackage[figuresright]{rotating}

\journal{Computer Vision and Image Understanding}

\begin{document}

\begin{frontmatter}



%% e.g. \dochead{17th International Conference on Dynamical Processes in Excited States of Solids}

\title{Water Detection through Spatio-Temporal Invariant Descriptors}

\author[label1]{Pascal Mettes\corref{cor1}\fnref{fn1}}
\ead{P.S.M.Mettes@uva.nl}
\author[label1]{Robby T. Tan\fnref{fn2}}
\ead{robbytan@unisim.edu.sg}
\author[label1]{Remco C. Veltkamp}
\ead{R.C.Veltkamp@uu.nl}

%\author{Pascal Mettes\corref{cor1}\fnref{fn1}}
%\ead{P.S.M.Mettes@uva.nl, +31634724306}
%\author{Robby T. Tan\fnref{fn2}}
%\ead{robbytan@unisim.edu.sg}
%\author{Remco C. Veltkamp}
%\ead{R.C.Veltkamp@uu.nl}

\address[label1]{Department of Information and Computing Sciences, Utrecht University, Utrecht, the Netherlands}

\cortext[cor1]{Corresponding author}

\fntext[fn1]{Present address: Intelligent Systems Lab Amsterdam, University of Amsterdam, Science Park 904, Amsterdam, the Netherlands}
\fntext[fn2]{Present address: Multimedia Technology and Design Programme, SIM University, Singapore}

\begin{abstract}
%%%%%%%%%%%%%%%%%% BELOW = VERSION BEFORE REVISION
%The detection of water in videos has a wide range of applications, yet the specific problem of water detection has mostly been ignored. Water detection has primarily been investigated in broader recognition problems, without distinctively investigating the properties of water. By not accounting for the invariance properties of water specifically, these methods are sub-optimal for water detection. Here, the behaviour of local water volumes is analysed. First, a video pre-processing algorithm is described, to increase invariance against water reflections and water colours. Second, a temporal descriptor and a spatial descriptor are advocated for locally detecting water. Third, a method is outlined for globally detecting water by means of local classification and global regularization. Experimental evaluation on the novel Video Water Database and the DynTex database indicates the effectiveness of the proposed algorithm, outperforming multiple algorithms from dynamic texture recognition and material recognition by ca. 5\% and 15\% respectively.
In this work, we aim to segment and detect water in videos. Water detection is beneficial for appllications such as video search, outdoor surveillance, and systems such as unmanned ground vehicles and unmanned aerial vehicles. The specific problem, however, is less discussed compared to general texture recognition. Here, we analyze several motion properties of water. First, we describe a video pre-processing step, to increase invariance against water reflections and water colours. Second, we investigate the temporal and spatial properties of water and derive corresponding local descriptors. The descriptors are used to locally classify the presence of water and a binary water detection mask is generated through spatio-temporal Markov Random Field regularization of the local classifications. Third, we introduce the Video Water Database, containing several hours of water and non-water videos, to validate our algorithm. Experimental evaluation on the Video Water Database and the DynTex database indicates the effectiveness of the proposed algorithm, outperforming multiple algorithms for dynamic texture recognition and material recognition by ca. 5\% and 15\% respectively.
\end{abstract}


\end{frontmatter}

% \linenumbers

\section{Introduction}
\begin{figure*}[t]
\begin{subfigure}{0.24\textwidth}
\includegraphics[width=\textwidth,height=0.1\textheight]{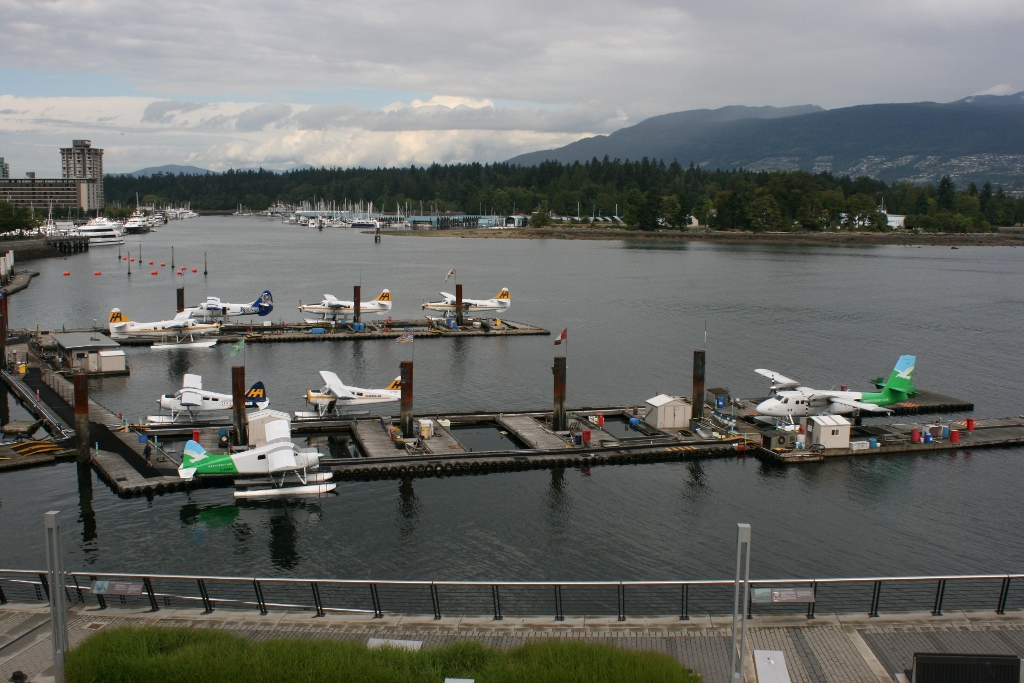}
\caption{Surveillance.}
\label{fig:app1}
\end{subfigure}
\begin{subfigure}{0.24\textwidth}
\includegraphics[width=\textwidth,height=0.1\textheight]{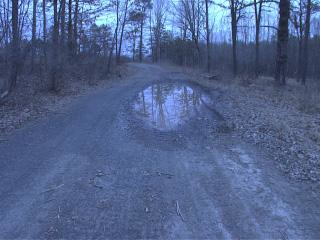}
\caption{Water hazards~\cite{ran06}.}
\label{fig:app2}
\end{subfigure}
\begin{subfigure}{0.24\textwidth}
\includegraphics[width=\textwidth,height=0.1\textheight]{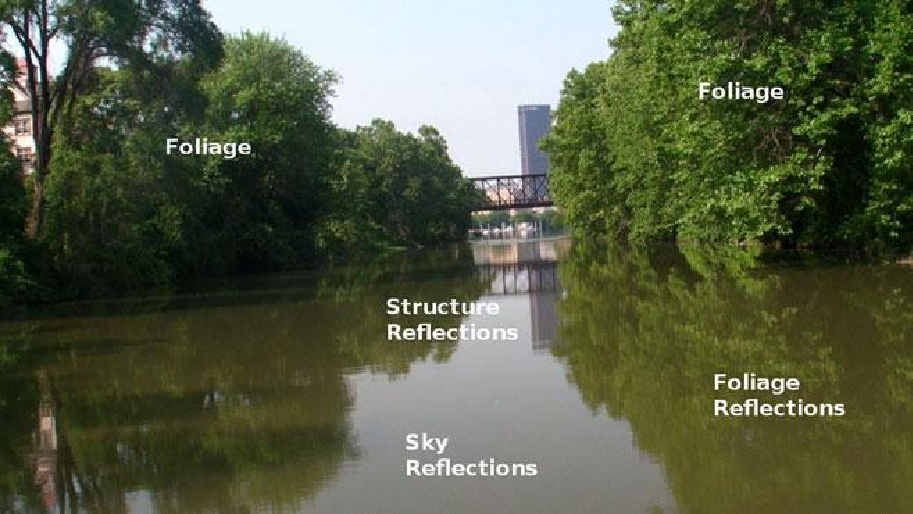}
\caption{Aerial water maps~\cite{sch12}.}
\label{fig:app3}
\end{subfigure}
\begin{subfigure}{0.24\textwidth}
\includegraphics[width=\textwidth,height=0.1\textheight]{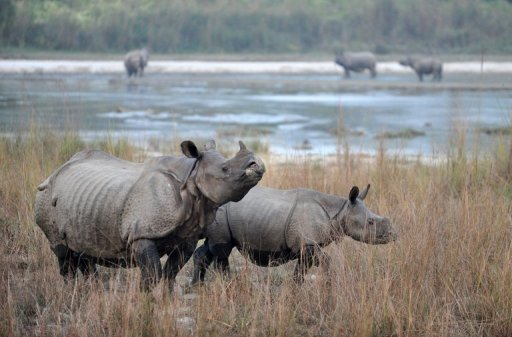}
\caption{Wildlife search~\cite{gem14}.}
\label{fig:app4}
\end{subfigure}
\caption{Visual examples of practical applications that benefit from water detection.}
\label{fig:applications}
\end{figure*}

The goal of this work is water detection in both natural and man-made environments from videos.
Spatio-temporal water detection finds applications in unmanned ground and aerial systems (e.g. self driving cars, and UAV's~\cite{gem14}), outdoor surveillance, video search, and wildlife search. These applications are highlighted in Figure~\ref{fig:applications}.
%We attest that there are a few important applications for water detection, particularly in unmanned ground and aerial systems. Self driving cards, for instance, need to detect water and avoid the water hazard. Other important applications, highlighted in Figure~\ref{fig:applications}, include outdoor surveillance, video search, and wildlife search.
To the best of our knowledge, related work focusses on texture recognition in general, and thus does not specifically explore the motion properties of water.

%We argue that multiple applications will benefit from water detection. Consider for example the applications shown in Fig.~1. The ability to detect water will aid for example surveillance in harbours (\ref{fig:app1}), water hazard detection in autonomous vehicles (\ref{fig:app2}), bird counting (\ref{fig:app3}), and wildlife search (\ref{fig:app4}). To our best knowledge, related work on static and dynamic texture recognition has however mostly ignored water detection as a separate problem. Here, we aim to fill this gap.

%The focus of this work is on investigating the invariance properties of water for the purpose of automatic spatio-temporal detection. In biological studies, the visual and physical properties of water have been investigated in order to understand the visual attractiveness of water in human and animal vision. From the work of Schwind~\cite{sch91}, it is known that water insects are attracted to the horizontal polarization caused by the reflections of water surfaces. This observation has for example been used to explain why certain insects lay eggs on highways~\cite{kri98}. In videos however, polarization information is not captured. Although it is possible to capture this information using polarization filters~\cite{hor97}, it is not practical for real-world scenario's. Furthermore, human observers are generally experts at detecting water in videos, indicating that water contains more visual properties that can be exploited for detection purposes. 

We focus on investigating the spatio-temporal motion properties of water. In biological studies, the visual properties of water have been investigated in order to understand the visual attractiveness of water in human and animal vision. From the work of Schwind~\cite{sch91}, it is known that water insects are attracted to the horizontal polarization caused by the reflections of water surfaces. This observation has for example been used to explain why certain insects lay eggs on highways~\cite{kri98}. In videos however, polarization information is not captured.
%Furthermore, human observers are generally experts at detecting water in videos, indicating that water contains other visual properties that can be investigated, as is done in this work.
Human observers are still experts at water detection without polarization information, indicating that water contains valuable spatio-temporal motion properties that can be exploited. Here, we investigate which spatio-temporal motion properties make water distinctive.

Current methods for automatic water detection can be divided into two categories: in specialized systems or as part of a broader recognition framework. In the broader fields of material recognition~\cite{hu11,met142,sha13} and dynamic texture recognition~\cite{cha08,dor03,zha072,faz09} water is one of the target classes. In these works, the objective is to minimize the miss-classification rate over all classes and as a result, the distinctive properties of water specifically are not investigated. Furthermore, the focus is generally on classification or segmentation, but not on the joint problem as posed here. On the other hand, water detection in specialized settings, such as autonomous driving \cite{ran06} and in maritime settings \cite{smi03}, either make non-generalizable restrictions on the movement and orientation of cameras \cite{ran06} or use auxiliary data sources in their measurements \cite{smi03,sch12,rat07}. To address the limitations of related work with respect to water detection specifically, this work provides an investigation into the temporal and spatial behaviour of water scenes.

This work reports three contributions. (1) We introduce a video pre-processing step to remove background reflections and inherent water colours. (2) We introduce a hybrid spatial and temporal descriptor for local water classification. For the temporal descriptor, we analyse the periodicity and regularity of local water patches and derive a descriptor that captures these elements. For the spatial descriptor, we
advocate Local Binary Patterns and we investigate what makes them suitable for local water detection. (3) We introduce a new dataset, the Video Water Database, for experimental evaluation and to encourage research into this topic. The Video Water Database, further discussed in section 5, along with the code used in the experimentation will be made publicly available to encourage further research into this topic.
%The goal of this work is to determine for each pixel in each frame in a video whether the pixel is part of a water surface. This work reports three contributions. First, a video pre-processing step is introduced to remove background reflections and inherent water colours. Second, a hybrid descriptor is introduced based on a fusion of a novel temporal descriptor and a spatial descriptor. Our temporal descriptor aims to incorporate three desired elements of motion related to water; gradual motion, repetitive motion, and regular motion. For the spatial descriptor, we advocate Local Binary Patterns and we investigate what makes them suitable for local water detection. Third, the Video Water Database is introduced for experimental evaluation and to encourage research into this topic. The Video Water Database, further discussed in section 5, along with the code used in the experimentation will be made publicly available to encourage further research into this topic.

%This work extends an earlier investigation into this topic~\cite{met141} in multiple aspects. First, the pre-processing stage is improved by exploiting Kernel Density Estimation to deal with areas on the border of multiple objects of reflection. Second, the temporal and spatial descriptors are further analyzed to indicate what makes these descriptors informative for water. Third, the experimental evaluation is extended by means of an evaluation of fusions of the proposed descriptors and additional qualitative results.

This work extends an earlier investigation into this topic~\cite{met141} in multiple aspects. An improvement is proposed in the pre-processing stage to deal with areas on the border of multiple objects of reflection, by modeling the density of pixel values over time. Also, further analysis is performed to investigate whether the hybrid descriptor is able to capture the spatial and temporal behaviour water ripples. In the experiments, we evaluate whether our method can generalize to water conditions and water types not seen during training. Lastly, another fusion of the temporal and spatial descriptor is evaluated.

The layout of the rest of this paper is as follows. In section 2, an overview of water detection in related work is provided. Section 3 introduces the pre-processing step of the videos and the analysis of the local behaviour of water. This is followed by the discussion on local probabilistic classification and spatio-temporal regularization in section 4. Finally, section 5 provides the experimental evaluation of the algorithm and the paper is concluded in section 6.

\section{Related work}
Given the lack of specific attention given to water detection, an overview is provided with respect to two broader recognition tasks: material recognition and dynamic texture recognition. Also, an overview of water localization in specialized systems is provided.

\subsection{Water in material recognition}

The classification of materials and static textures in images has a long history of investigation~\cite{sha13, eve12, oja02, var09}. Works on this topic are in line with Biederman~\cite{bie87}, who conjectured that materials are recognized in human vision by their surface characteristics such as texture and colour. Well-known exemplary approaches include the use of filter bank distributions \cite{var05}, Local Binary Patterns \cite{oja02}, and image patch exemplars \cite{var09}. In recent works, a shift has been made from image databases made in a laboratory setting~\cite{oja02,var05,var09} to real-world image databases~\cite{hu11, met142, sha13}. In these works, a range of surface characteristics, e.g.\ texture, colour, reflectance, and curvature, is extracted to find out what characteristics are best for classification. The results of these works indicate that spatial information is informative for distinguishing different materials. For water detection in videos however, there are two limiting aspects. First, only the spatial characteristics are investigated, excluding valuable temporal information. Second, research into material recognition has focused mostly on solving the classification problem or the segmentation problem, but not their joint problem.

\subsection{Water in dynamic texture recognition}
Dynamic textures are part of a class of motions with either structural or statistical similarity in both space and time \cite{nel92}. Exemplary dynamic textures include fire, water, flags, and weather patterns. One of the dominant approaches in dynamic texture recognition is based on optical flow statistics~\cite{faz09, vid05, faz07, che13}. In these approaches, either a global description is generated using invariant flow statistics such as characteristic direction and magnitude of flow vectors~\cite{faz07}, or flow vectors are binned into Histograms of Optical Flow (HOOF)~\cite{che13}. The use of optical flow is intuitively interesting for water detection, as the spatio-temporal movement of water seems statistically different to related textures. The use of conventional optical flow is however problematic for water detection in videos, as water meets none of the requirements for a proper flow estimation: Lambertian surface reflectance, pure translational motion parallel to the image plane, and uniform illumination~\cite{bea95}. A representation using optical flow will therefore be heavily influenced by the noise of the flow estimation, which makes optical flow not desirable for water detection.

\begin{figure*}[t]
\centering
\resizebox{\textwidth}{!}{
\begin{tikzpicture}
\node[anchor=south west,inner sep=0] at (0,2.375) {\includegraphics[width=0.205\textwidth]{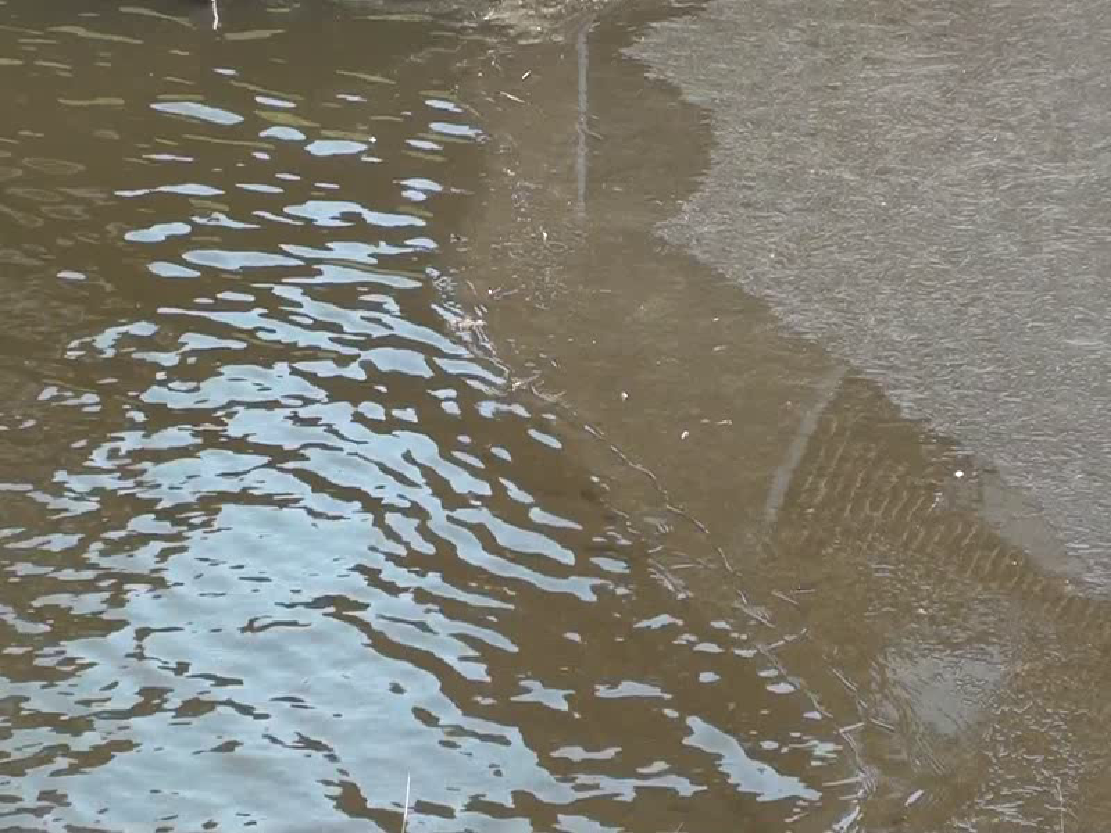}};
\draw [draw=red, line width=0.4mm] (0.3,2.7) rectangle (1.4,3.6);
\draw [->] (4,3.75) -- (6.5,5) node [midway, above, sloped] {direct mode};
\node[anchor=south west,inner sep=0] at (6.75,3.75) {\includegraphics[width=0.205\textwidth]{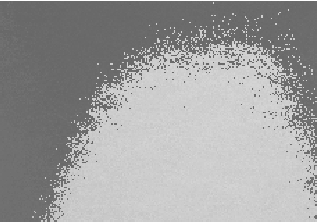}};
\draw [->] (10.75,5) -- (13.75,5)  node [midway, above] {mode subtraction};
\node[anchor=south west,inner sep=0] at (14,3.75) {\includegraphics[width=0.205\textwidth]{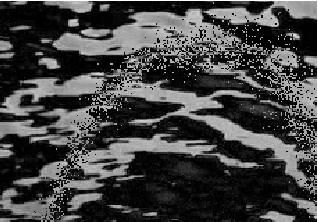}};
\draw [->] (4,3.75) -- (6.5,2.25) node [midway, below, sloped] {density mode};
\node[anchor=south west,inner sep=0] at (6.75,1) {\includegraphics[width=0.205\textwidth]{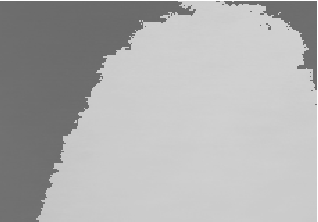}};
\draw [->] (10.75,2.25) -- (13.75,2.25)  node [midway, above] {mode subtraction};
\node[anchor=south west,inner sep=0] at (14,1) {\includegraphics[width=0.205\textwidth]{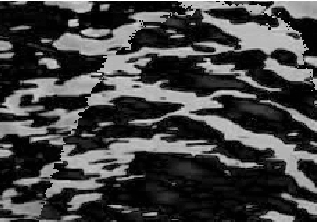}};
\end{tikzpicture}}
\caption{The process of removing reflection highlights and water colours. Left: frame of a video containing water. Top middle and right: the temporal mode computed using Eq.~\ref{eq:mode1} over the whole video and the corresponding residual of the frame, both for the selected red region. Bottom middle and right: the temporal mode computed using Eq.~\ref{eq:mode2} and the corresponding residual.}
\label{fig:mode}
\end{figure*}
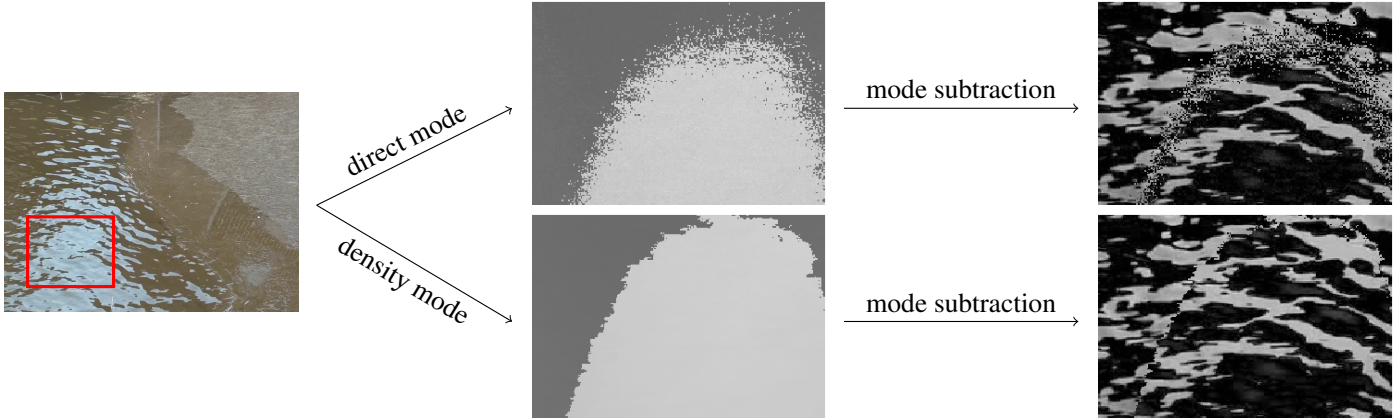

Another popular research direction focuses on modeling dynamic textures as Linear Dynamical Systems (LDS)~\cite{cha08, dor03, sai01, rav13}. In dynamic texture recognition, the use of LDS has been made popular by Saison et al.~\cite{sai01} and Doretto et al.~\cite{dor03}, mostly due to the proposed efficient sub-optimal learning procedure. As the original formulation of LDS requires a modeling of whole videos, it is unfit for local detection purposes. In order to deal with multiple textures within a video, several extensions have been provided. These include mixtures of dynamic textures~\cite{cha08}, hierarchical EM clustering~\cite{mum13}, and Bags of Dynamical Systems~\cite{rav13}. These algorithms can potentially handle multiple textures in a video, but they have so far not been applied to detection problems. A noteworthy exception is the work of Ravichandran et al.~\cite{rav11}, where the joint segmentation and classification problem of dynamic textures is tackled by dividing a video into parts using Dynamic Appearance Images computed from LDS, after which the parts are represented by a bag-of-words representation with SIFT features. The representations are however more general and not tailored to water detection. Also, it is explicitly assumed that the texture class of a pixel does not change over time, restricting potential applications. Rather than performing a holistic modeling as with LDS, this work attempts to detect water from a local scale. The local scale is essential, as water is not bound to specific shapes in a scene.

Notable is also the research on spatio-temporal Local Binary Patterns for dynamic texture classification~\cite{zha07}. In the work of Zhao and Pietik\"{a}inen~\cite{zha07}, both Volume LBP (VLBP) and LBP-TOP are introduced. VLBP generates longer histograms by simply adding binary comparison to temporal neighbours. Since the length of the histogram increases exponentially with the number of comparisons, VLBP typically yields histograms the size of $2^{14}$ or $2^{26}$. More compact representations can be generated with LBP-TOP, but still this leads to longer representations than the purely spatial variant~\cite{oja02}. The length of the feature representation limits the applicability of VLBP and LBP-TOP for local detection, but the lower dimensional spatial LBP remains interesting for water detection.

\subsection{Water localization in specialized systems}
Water detection has been investigated in specialized systems, including autoonomous driving systems~\cite{ran06,ran10,iqb09}, maritime environments~\cite{smi03}, and using flying robots~\cite{sch12,rat07}. Although these algorithms might provide a suitable solution in their restricted environment, none of the mentioned works are able to generalize to fully automatic water detection using minimally constrained video material.

In autonomous driving, several works have attempted to detect water hazard such as puddles and canals, in order to inform the autonomous agent. In the work of Rankin and Matthies~\cite{ran06}, colour and texture cues are combined with stereo information to indicate water regions. Furthermore, estimated elevations are used to detect ground regions, in order to decrease the false positive rate. A similar method is introduced in~\cite{iqb09}. A subsequent performance evaluation by Rankin et al.~\cite{ran10} is also focused on the more specific scenario where stereo information is provided. In all works, additional sensors are used to help the detection problem. In both maritime settings and in works using flying robots, similar non-generalizable assumptions have been made, whether it is assuming that water is within a specific part of the frame~\cite{smi03}, requires a manual pre-processing step to identify sky regions~\cite{sch12}, or uses auxiliary sensors~\cite{sch12,rat07}. The works do therefore not generalize to water detection with minimal camera assumptions and without additional sensors, rendering them impractical for the problem of this work.

\section{Local spatio-temporal water analysis}
%Due to the deformable nature of water, it is not bound to a specific shape~\cite{bie87}. As such, generating a detection algorithm based on the global shape of a water surface, as is done in object detection \cite{fel10,wan13}, is an inappropriate approach. This work is therefore focused on the identification of water on a local scale. By examining and classifying local patches and volumes independently, the deformable shapes of water can be handled.
%A detection algorithm based on the global shape of a water surface, as is done in object detection \cite{fel10,wan13}, is inappropriate, as water is not bound to a specific shape~\cite{bie87}. This work is therefore focused on the identification of water on a local scale. By examining and classifying local patches and volumes independently, the deformable shape of water can be handled.

Since natural water scenes are dominated by aspects such as water colours, sky reflection, and object reflections, the videos are first pre-processed in a single-pass offline process. The pre-processing of a video results in a residual video, where these aspects are removed, to focus solely on water waves and ripples. After that, the temporal and spatial behaviour of water is analysed, resulting in a novel temporal descriptor and the use of Local Binary Patterns \cite{oja02,qia11} as a spatial descriptor. By combining these descriptors into a hybrid descriptor, it becomes possible to locally detect water.

\begin{figure*}[t]
\centering
\resizebox{\textwidth}{!}{
\begin{tikzpicture}
\node[inner sep=0] at (-1,2.5) {\includegraphics[width=1.5cm]{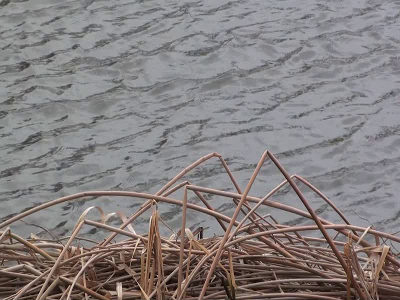}};
\draw [draw=blue, line width=0.5mm] (-1.75,1.95) rectangle (-0.25,3.05);
\node[inner sep=0] at (-1,1.25) {\includegraphics[width=1.5cm]{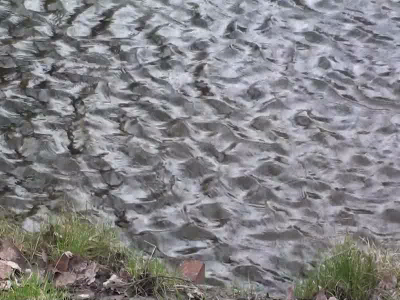}};
\draw [draw=blue, line width=0.5mm] (-1.75,0.7) rectangle (-0.25,1.8);
\node[inner sep=0] at (-1,0) {\includegraphics[width=1.5cm]{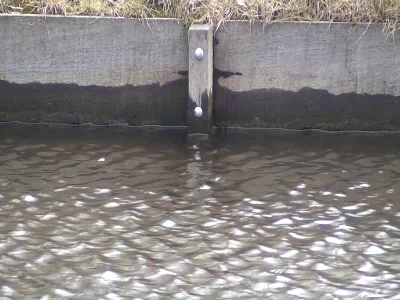}};
\draw [draw=blue, line width=0.5mm] (-1.75,-0.55) rectangle (-0.25,0.55);

\node[inner sep=0] at (1.5,2.5) {\includegraphics[width=1.5cm]{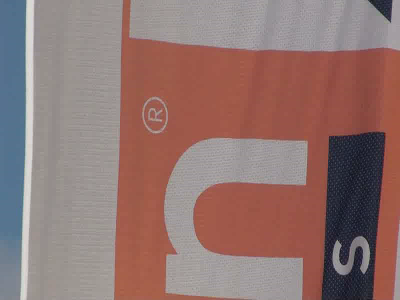}};
\draw [draw=red, line width=0.5mm] (0.75,1.95) rectangle (2.25,3.05);
\node[inner sep=0] at (1.5,1.25) {\includegraphics[width=1.5cm]{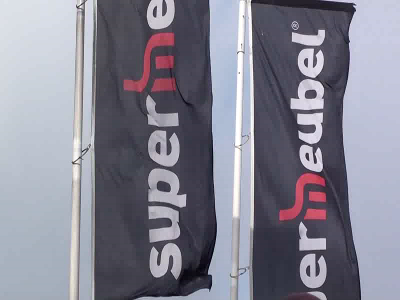}};
\draw [draw=red, line width=0.5mm] (0.75,0.7) rectangle (2.25,1.8);
\node[inner sep=0] at (1.5,0) {\includegraphics[width=1.5cm]{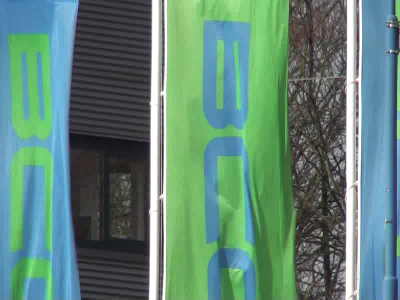}};
\draw [draw=red, line width=0.5mm] (0.75,-0.55) rectangle (2.25,0.55);

\draw [->] (2.7,1.25) -- (3.9,1.25)  node [midway, above] {Signal};

\node[inner sep=0] at (6.25,1.25) {\includegraphics[width=4cm]{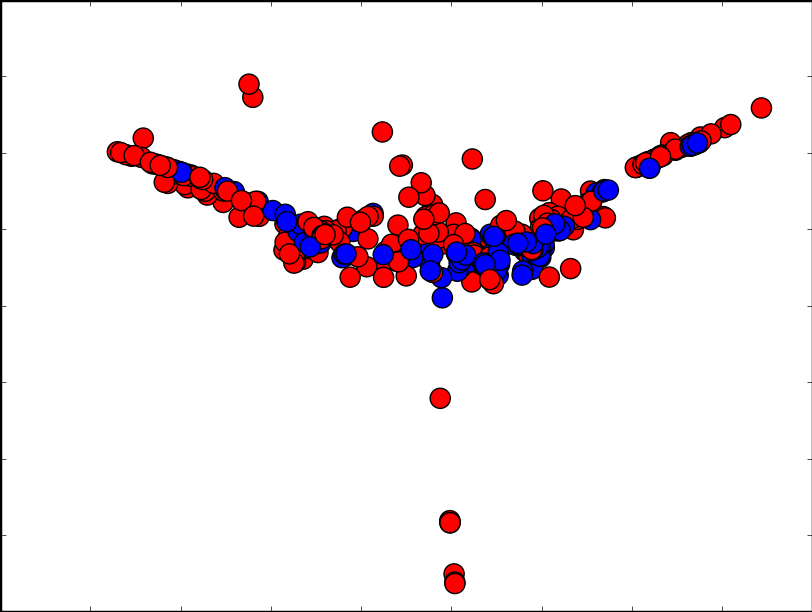}};
\draw [draw=black, line width=0.5mm] (4.25,-0.25) rectangle (8.25,2.75);

\draw [->] (8.6,1.25) -- (9.8,1.25)  node [midway, above] {$\ell_{1}$ FFT};

\node[inner sep=0] at (12.15,1.25) {\includegraphics[width=4cm]{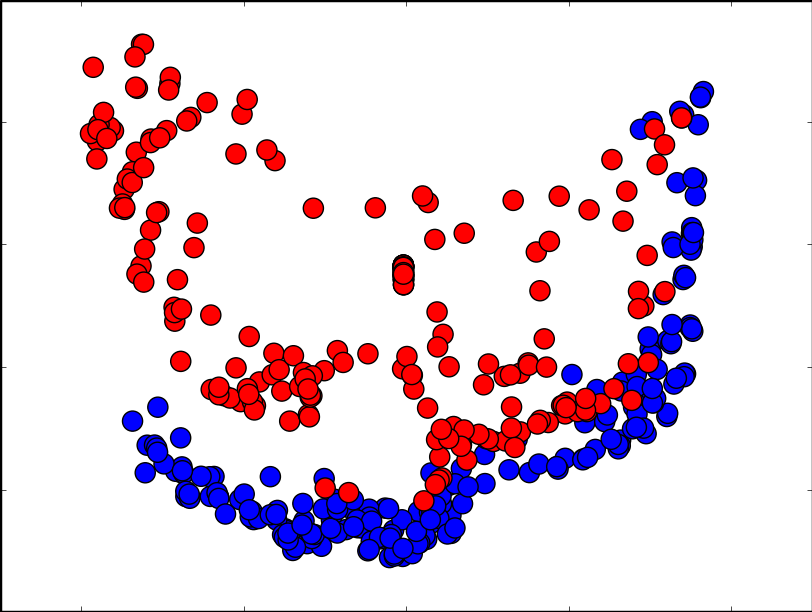}};
\draw [draw=black, line width=0.5mm] (10.15,-0.25) rectangle (14.15,2.75);
\end{tikzpicture}}
\caption{Illustration of the effect of the proposed signal transformations. Left: a frame is shown for 3 water (blue) and 3 non-water (red) videos. Middle: a 2D projection of sampled local signals is shown for the water and non-water videos. Right: a 2D projection of the same signals after the transformations. Note that the signals of the water and non-water videos become nicely separated after the transformations, even in a 2D projection.}
\label{fig:signal}
\end{figure*}
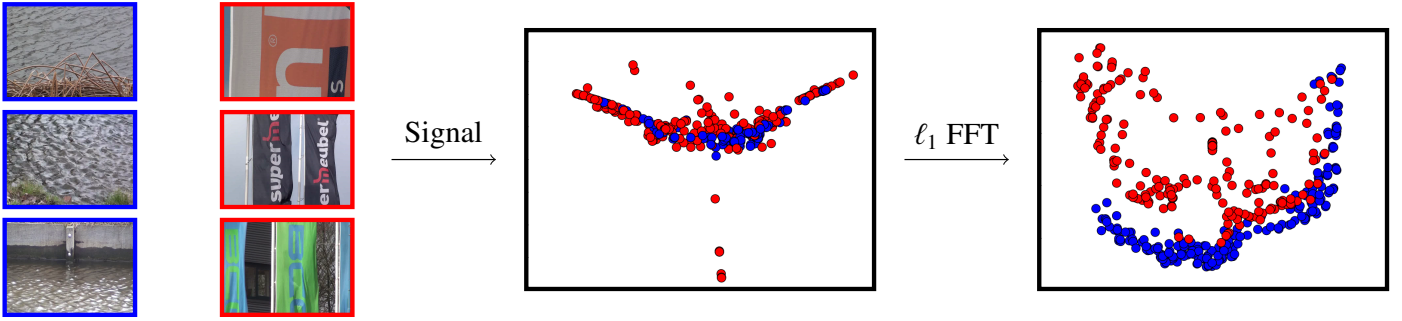

\subsection{Residual videos}
An important part in the process of detecting natural water scenes is dealing with the inherent variability of water, due to water colour, reflections, ripples, waves, and weather conditions. Instead of exploiting consistencies among these elements of variability, the focus of this work is to generate descriptions that are invariant to these variations.

To accommodate the temporal and spatial descriptor towards a distinctive water representation that is invariant to elements such as reflections and water colours, the videos in the database are first pre-processed. The goal of this step is to capture and subtract water reflections and colours from the video frames, yielding residual frames. The dominant reflections and colours are obtained for each pixel by computing the mode value over the frames. The underlying assumption is that water ripples and waves form a temporary disruption of an otherwise direct reflection. The most often occurring intensity values indicate the dominant water colours and reflections. As such, the temporal mode frame $M$ of a video is defined pixel-wise as follows:
\begin{equation}
M(x,y) = \argmax_{i} \sum_{j=1}^{t} [\![I_{j}(x,y) = i]\!],
\label{eq:mode1}
\end{equation}
where $t$ denotes the number of frames in the video, the Iverson brackets $[\![\cdot]\!]$ denote the indicator function, and $i \in \{0,..,255\}$ denotes the set of intensity values. The residual frames can be obtained simply by means of absolute differencing the frames of the video with the temporal mode frame.

The use of the temporal mode for each pixel is not a stable choice under all circumstances. Most notably, as illustrated in Fig.~\ref{fig:mode}, on strong edges, e.g.\ on the border of sky and object reflections, this approach yields a noisy result. To create more coherent residuals, Kernel Density Estimation \cite{sco09} is performed over the set of intensity values for each pixel. In other words, for a pixel $(x,y)$, the mode value is determined as:
\begin{equation}
M(x,y) = \argmax_{i} \frac{1}{t} \sum_{j=1}^{t} K_{h}(i - I_{t}(x,y)),
\label{eq:mode2}
\end{equation}
where $K_{h}(\cdot)$ denotes the Gaussian kernel and the bandwidth $h$ is estimated using Scott's Rule \cite{sco09}. In Fig.~\ref{fig:mode}, the mode frame using KDE is also shown for the example video. Contrary to the original method, this approach yields proper mode frames even at boundaries of two or more dominant colours.

The frames of the videos are downsized to a quarter of their width and height, in order to remain computationally practical. This is since the use of Gaussian KDE over Eq. \ref{eq:mode2} increases the time complexity of computing the mode frame of a video from $O(p(t + i))$ to $O(pti)$, with $p$ the number of pixels, $t$ the number of frames, and $i$ the number of intensity values.

\subsection{Local temporal water behaviour}
%With the water waves and ripples highlighted by the pre-processing step outlined above, the aim here is to investigate both the temporal and spatial behaviour of water for purely discriminative purposes. In the first analysis, the temporal behaviour is investigated, while the spatial behaviour is investigated in the next Section. The separate descriptors derived from the analyses can be combined using early or late fusion~\cite{sno05}.

For the temporal descriptor, a Eulerian approach is opted over a Lagrangian. In other words, rather than tracking pixels over time as is done with optical flow (the Lagrangian approach), the dynamics of water is investigated from static locations. The hypothesis behind this is that transitions of brightness values over time contain valuable information regarding the characteristics of water. It is hypothesized that they include gradual motion (waves enter and exit a local area smoothly), repetitive motion (waves re-occur in similar fashion over time), and regular motion (waves re-occur at similar intervals).

As the brightness transitions of individual pixels are sensitive to noise, the local temporal behaviour of water is analysed by averaging brightness values of a local region around a pixel. For a spatio-temporal video volume, an $m$-dimensional signal is generated by computing the mean brightness value of an $n \times n$ patch around a pixel for $m$ consecutive frames. Note that this is similar to a 3D mean convolution filter of size $n \times n \times m$. The resulting list of brightness values can be seen as a signal. These signals exhibit more sinusoidal patterns when extracted locally from water regions than from non-water regions, as is expected from the hypothesized motion characteristics of water.

Using the signals obtained from the local 3D convolution, the primary concern becomes finding a descriptor that generates a small distance between two water signals and a large distance between a water and non-water signal with respect to the hypotheses. An obvious solution is to directly compute the $\ell_{2}$ distance between two signals, i.e.\ to directly use the signals as the temporal water descriptor. This solution is however erroneous, as the signals lack a number of invariance properties. A descriptor based on the $m$-dimensional signals should in effect be invariant to temporal shifts, brightness shifts, and brightness amplitudes. This can be generated by computing the minimum distance between two signals $S_{1}$ and $S_{2}$ under all temporal shifts $T$, brightness shifts $B$, and amplitudes $A$:
\begin{equation}
d(S_{1}, S_{2}) = \min_{t \in T, b \in B, a \in A} \sum_{i=1}^{m} S_{1}[i] - a \cdot (S_{2}[i + t] + b).
\end{equation}
The above equation is however prohibitively expensive. A more scalable approach is to create temporal and brightness shift invariance by computing the Fourier Transform. For an $m$-dimensional signal $S$, the $m$-dimensional Fourier transform $F$ is computed as follows:
\begin{equation}
F_{i} = | \sum_{j=1}^{m} S_{j} \exp(-2\pi i j \sqrt{-1} m) |.
\label{eq:fft1}
\end{equation}
Note that in Eq.~\ref{eq:fft1}, the variable $i$ does not denote the imaginary number, but the index of the Fourier transform; the imaginary number is for convenience written explicitly as $\sqrt{-1}$.

Computing distances between two Fourier signals creates (temporal and brightness) shift invariance in $O(m \log m)$ time. However, since the descriptor is not invariant against amplitudes, the final temporal descriptor is generated by performing $\ell_{1}$ normalization. An invariance with respect to brightness amplitudes is desirable, as two descriptors with similar levels of regularity and repetition will have a large distance in both the original signal space and the Fourier transform space if their amplitudes are not similar (e.g. rough and calm water). Using the $\ell_{1}$ normalization to add the final layer of invariance, a temporal water descriptor $\{F_{i}\}_{i=1}^{m}$ is computed from an original signal $\{S_{i}\}_{i=1}^{m}$ as:
\begin{equation}
F_{i} = \frac{| \sum_{j=1}^{m} S_{j} \exp(-2\pi i j \sqrt{-1} m) |}{\sum_{k=1}^{m} | \sum_{j=1}^{m} S_{j} \exp(-2\pi k j \sqrt{-1} m) |}.
\label{eq:fft2}
\end{equation}
A practical justification of adding the layers of invariance is provided in Fig~\ref{fig:signal}. In the example of the Figure, signals are randomly sampled from 3 water and 3 flag videos. In the 2D projection~\cite{ten00} of the original signals, the water and flag signals are completely indistinguishable. After adding the desired elements of invariance, the signals become nicely separable, even in a 2D projection of 200D descriptors.

\subsection{Local spatial water behaviour}
Although the above introduced water descriptor can capture the temporal behaviour of water, it explicitly ignores the spatial layout of water waves and ripples. Due to the deformable nature of water, a descriptor is desired that can provide spatial information without explicitly modeling water waves and ripples. To meet this desire, Local Binary Pattern histograms~\cite{oja02,qia11} are investigated. LBP histograms have a number of benefits particularly desired properties for the purpose of this work. First and foremost, the spatial arrangement of individual pixels only extends to a one pixel neighbourhood. This is convenient, because of the deformations possible within a patch. On the other hand, the histograms are of sufficient dimensionality to be discriminative.

\begin{figure}[t]
\centering
\begin{subfigure}[b]{\linewidth}
\centering
\resizebox{\textwidth}{!}{
\begin{tikzpicture}
\draw [fill=white] (0,0) rectangle (0.5,0.5); \draw [fill=gray] (0,0.5) rectangle (0.5,1); \draw [fill=gray] (0,1) rectangle (0.5,1.5);
\draw [fill=white] (0.5,0) rectangle (1,0.5); \draw [fill=white] (0.5,0.5) rectangle (1,1); \draw [fill=gray] (0.5,1) rectangle (1,1.5);
\draw [fill=white] (1,0) rectangle (1.5,0.5); \draw [fill=gray] (1,0.5) rectangle (1.5,1); \draw [fill=gray] (1,1) rectangle (1.5,1.5);
\draw (0.5,0.5) -- (1,1); \draw (1,0.5) -- (0.5,1);

\draw [fill=gray] (2,0) rectangle (2.5,0.5); \draw [fill=gray] (2,0.5) rectangle (2.5,1); \draw [fill=white] (2,1) rectangle (2.5,1.5);
\draw [fill=gray] (2.5,0) rectangle (3,0.5); \draw [fill=white] (2.5,0.5) rectangle (3,1); \draw [fill=white] (2.5,1) rectangle (3,1.5);
\draw [fill=gray] (3,0) rectangle (3.5,0.5); \draw [fill=white] (3,0.5) rectangle (3.5,1); \draw [fill=white] (3,1) rectangle (3.5,1.5);
\draw (2.5,0.5) -- (3,1); \draw (2.5,1) -- (3,0.5);

\draw [fill=white] (4,0) rectangle (4.5,0.5); \draw [fill=gray] (4,0.5) rectangle (4.5,1); \draw [fill=gray] (4,1) rectangle (4.5,1.5);
\draw [fill=white] (4.5,0) rectangle (5,0.5); \draw [fill=white] (4.5,0.5) rectangle (5,1); \draw [fill=gray] (4.5,1) rectangle (5,1.5);
\draw [fill=white] (5,0) rectangle (5.5,0.5); \draw [fill=white] (5,0.5) rectangle (5.5,1); \draw [fill=gray] (5,1) rectangle (5.5,1.5);
\draw (4.5,0.5) -- (5,1); \draw (4.5,1) -- (5,0.5);

\draw [fill=white] (6,0) rectangle (6.5,0.5); \draw [fill=white] (6,0.5) rectangle (6.5,1); \draw [fill=white] (6,1) rectangle (6.5,1.5);
\draw [fill=gray] (6.5,0) rectangle (7,0.5); \draw [fill=white] (6.5,0.5) rectangle (7,1); \draw [fill=white] (6.5,1) rectangle (7,1.5);
\draw [fill=gray] (7,0) rectangle (7.5,0.5); \draw [fill=gray] (7,0.5) rectangle (7.5,1); \draw [fill=white] (7,1) rectangle (7.5,1.5);
\draw (6.5,0.5) -- (7,1); \draw (6.5,1) -- (7,0.5);
\end{tikzpicture}}
\caption{}
\end{subfigure}
\vspace{0.05cm}\\
\begin{subfigure}[b]{\linewidth}
\centering
\resizebox{\textwidth}{!}{
\begin{tikzpicture}
\draw [fill=white] (0,0) rectangle (0.5,0.5); \draw [fill=white] (0,0.5) rectangle (0.5,1); \draw [fill=white] (0,1) rectangle (0.5,1.5);
\draw [fill=white] (0.5,0) rectangle (1,0.5); \draw [fill=white] (0.5,0.5) rectangle (1,1); \draw [fill=white] (0.5,1) rectangle (1,1.5);
\draw [fill=gray] (1,0) rectangle (1.5,0.5); \draw [fill=gray] (1,0.5) rectangle (1.5,1); \draw [fill=gray] (1,1) rectangle (1.5,1.5);
\draw (0.5,0.5) -- (1,1); \draw (1,0.5) -- (0.5,1);

\draw [fill=gray] (2,0) rectangle (2.5,0.5); \draw [fill=gray] (2,0.5) rectangle (2.5,1); \draw [fill=gray] (2,1) rectangle (2.5,1.5);
\draw [fill=gray] (2.5,0) rectangle (3,0.5); \draw [fill=white] (2.5,0.5) rectangle (3,1); \draw [fill=gray] (2.5,1) rectangle (3,1.5);
\draw [fill=gray] (3,0) rectangle (3.5,0.5); \draw [fill=white] (3,0.5) rectangle (3.5,1); \draw [fill=white] (3,1) rectangle (3.5,1.5);
\draw (2.5,0.5) -- (3,1); \draw (2.5,1) -- (3,0.5);

\draw [fill=gray] (4,0) rectangle (4.5,0.5); \draw [fill=gray] (4,0.5) rectangle (4.5,1); \draw [fill=gray] (4,1) rectangle (4.5,1.5);
\draw [fill=white] (4.5,0) rectangle (5,0.5); \draw [fill=white] (4.5,0.5) rectangle (5,1); \draw [fill=gray] (4.5,1) rectangle (5,1.5);
\draw [fill=gray] (5,0) rectangle (5.5,0.5); \draw [fill=gray] (5,0.5) rectangle (5.5,1); \draw [fill=gray] (5,1) rectangle (5.5,1.5);
\draw (4.5,0.5) -- (5,1); \draw (4.5,1) -- (5,0.5);

\draw [fill=white] (6,0) rectangle (6.5,0.5); \draw [fill=white] (6,0.5) rectangle (6.5,1); \draw [fill=white] (6,1) rectangle (6.5,1.5);
\draw [fill=white] (6.5,0) rectangle (7,0.5); \draw [fill=white] (6.5,0.5) rectangle (7,1); \draw [fill=gray] (6.5,1) rectangle (7,1.5);
\draw [fill=gray] (7,0) rectangle (7.5,0.5); \draw [fill=gray] (7,0.5) rectangle (7.5,1); \draw [fill=white] (7,1) rectangle (7.5,1.5);
\draw (6.5,0.5) -- (7,1); \draw (6.5,1) -- (7,0.5);
\end{tikzpicture}}
\caption{}
\label{fig:lbpnegcorr}
\end{subfigure}
\caption{Illustration of LBP values that (a) correlate positively to water and (b) correlate negatively to water. White boxes indicate a value of 1. Note how the positive LBP values are smooth, while the negative LBP values are not as smooth and can also have gaps.}
\label{fig:lbpcorr}
\end{figure}
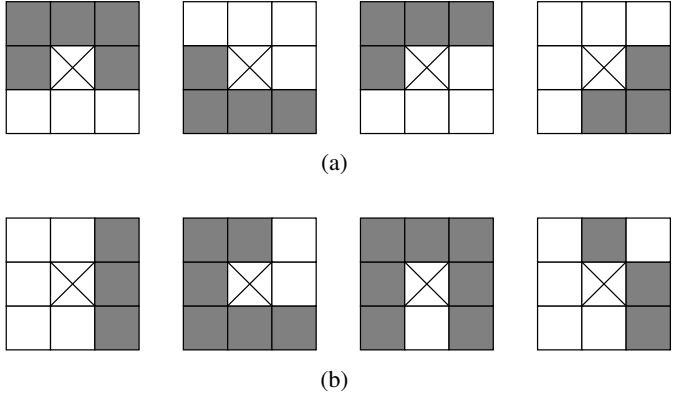

The LBP value of a single pixel is computed by comparing the intensity value of the pixel to nearby pixels. Here, the LBP-variant using the 8 direct neighbours is explored, i.e.\ the LBP-value of a pixel is determined as:
\begin{equation}
\mbox{LBP}(g^{c}) = \sum_{p=0}^{7} [\![g_{p}^{c} - g^{c} \geq 0]\!] \cdot 2^{p},
\label{eq:lbp}
\end{equation}
where $g^{c}$ denotes the pixel to be evaluated and $\{g_{p}^{c}\}_{p=0}^{7}$ denotes the 8 direct neighbours. In order to compute a descriptor over a local region, the LBP-value is computed for each pixel in that region and placed in one of the $2^{8} = 256$ bins, according to the value yielded by Eq. \ref{eq:lbp}.

The question remains whether the use of LBP histograms is beneficial for water specifically. To investigate this, a number of local patches of water and non-water videos are randomly extracted and trained using a linear classifier, in this case a linear Support Vector Machine. In Fig.~\ref{fig:lbpcorr}, both positively and negatively correlated Local Binary Pattern values are shown as a function of the coefficients of the trained Support Vector Machine. From that Figure, it can readily be observed that a LBP classifier fires on the edges of ripples. Furthermore, it has been observed in the analysis of the SVM classifier that only a third of the LBP values are positively correlated at all with water. A substantial part of the LBP values with more than two transitions (similar to the rightmost example of Fig.~\ref{fig:lbpnegcorr}) are to some extend negatively correlated to water, indicating that the spatial layout of water waves and ripples is not chaotic and is characterized by smooth spatial transitions.

Note that the use of Local Binary Patterns results in a descriptor aimed at extracting gradient information. Throughout this work, it is however referred to as a spatial descriptor to exemplify the contrast to the temporal descriptor; the temporal descriptor tries to identify patterns in the temporal dimension, the spatial descriptor does the same in the two spatial dimensions.

\section{Classification and regularization} \label{sec:classreg}
Given the pre-processed videos and a local temporal and spatial descriptor, the final goal becomes generating a detection mask for each frame of a video. This is performed in two steps; direct probabilistic classification and spatio-temporal regularization. In the first step, a model is created from positively and negatively sampled descriptors. This model can then be applied to a test video, resulting in a large number of independent classifications. These classifications can already be served as detections by binarizing all the probabilities. As the learned model does not perfectly classify each local video volume, the output of the individual classifications is noisy and a form of regularization is required to generate coherent water detection masks.

The derived temporal and spatial descriptors are not used as local features for a more global encoding; rather, a model is generated directly from individual descriptors. In this work, both the early and late fusion variant are experimentally evaluated~\cite{sno05}. In early fusion, the temporal and spatial descriptors computed from a local video volume are first concatenated, after which a model is trained on these hybrid descriptors. Contrarily, in late fusion, a model is trained separately for the temporal and spatial descriptors, and the probability of a local video volume of being water is determined by averaging the scores from the two models.

\subsection{Local probabilistic classification} \label{sec:class}
Classification is performed by sampling local descriptors from training videos. These descriptors are then used as feature vectors for the training of the classifier, where the labels are inherited from the video from which they were sampled. As the total number of local video patches and volumes over all training videos is cumbersomely large, a random sampling approach is adopted here. To maximize coverage, each frame of each training video is evaluated, and a low number of descriptors are extracted from randomly sampled locations.

The yielded set of feature vectors are then fed to a Random Decision Forest~\cite{cri12}. The use of a Decision Forest is particularly interesting, since it provides probabilistic outputs and it inherently generates non-linear decision boundaries. Probabilistic outputs will prove to be useful, as the uncertainty can be used for regularizing the detection. Local descriptors are extracted and independently classified for each frame of a test video.

\subsection{Spatio-temporal regularization}
The procedure of Section~\ref{sec:class} generates hundreds of individual local water probabilities per frame of a test video. As the classification procedure is not expected to be fully accurate, a number of miss-classifications are bound to occur, even within a single frame. The additional information gained by computing probabilities over binary labels opens up the possibility to handle classification outliers. Under the intuitive assumption that water regions have a high spatial support (i.e.\ there are a limited number of boundaries between water and non-water regions), a final detection map per frame of a video can be computed by means of regularization. Here, the regularization takes of form of a binary Markov Random Field~\cite{boy04}, that attempts to solve the following minimization objective:
\begin{equation}
f(x,y) = \sum_{p \in V} V_{p}(x_{p}) + \lambda \sum_{(p,q) \in C} V_{pq}(x_{p}, x_{q}),
\label{eq:mrf}
\end{equation}
where the unitary term $V_{p}$ denotes the match between the label of node $p$ and its corresponding probability:
\begin{equation}
V_{p}(x_{p}) = \left\{ \begin{array}{l l}
                    1 - P_{p} & \quad \mbox{if $x_{p}$ is water}\\
                    P_{p} & \quad \mbox{otherwise,}
               \end{array} \right.
\end{equation}
where $P_{p}$ denotes the probability of node $p$ of being water. The pairwise term $V_{pq}$ of Eq.~\ref{eq:mrf} follows the well-known $0/1$ Potts model that enforces similarity between the labels of nodes from the same clique~\cite{boy04}. The term $\lambda$ is a hyperparameter weighting the importance between the unitary and pairwise terms.

\begin{figure*}[t]
\centering
\includegraphics[width=0.16\textwidth]{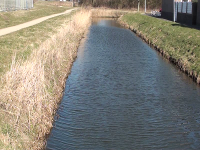}
\includegraphics[width=0.16\textwidth]{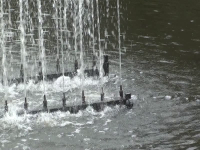}
\includegraphics[width=0.16\textwidth]{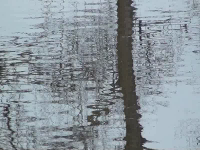}
\includegraphics[width=0.16\textwidth]{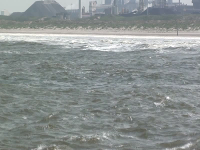}
\includegraphics[width=0.16\textwidth]{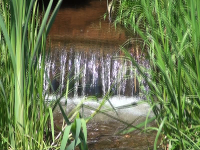}
\includegraphics[width=0.16\textwidth]{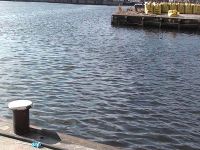}\\
\hspace{0.001cm}\\
\includegraphics[width=0.16\textwidth]{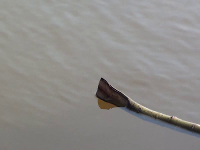}
\includegraphics[width=0.16\textwidth]{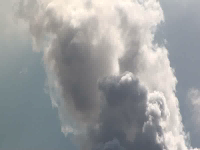}
\includegraphics[width=0.16\textwidth]{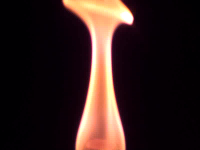}
\includegraphics[width=0.16\textwidth]{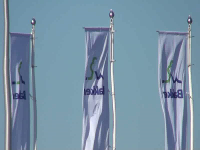}
\includegraphics[width=0.16\textwidth]{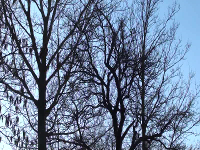}
\includegraphics[width=0.16\textwidth]{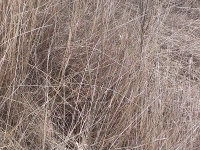}
\caption{An example frame of each of the subcategories present in the Video Water Database. See text for details on the subcategories.}
\label{fig:vwd}
\end{figure*}

An obvious choice of cliques in the Markov Random Field are the pairwise spatial neighbours within a single frame of a video. This would involve generating a single Markov Random Field for each frame. As it is furthermore desired to penalize different labelings at the same location between consecutive frames, the pairwise temporal neighbours are also used as cliques. This results in a single spatio-temporal Markov Random Field for each evaluated video.

\section{Experimentation}
To validate the proposed algorithm for water detection, the algorithm is evaluated on two different but related tasks; water detection and water classification-by-selection.

The detection quality of a video is defined as the average of the fit of the detection fit per frame. Formally, the detection fit $D$ of a binarized video $V$ compared to a ground truth mask $M$ is defined as:
\begin{equation}
D(V,M) = \frac{\sum_{i=1}^{|V|} d(V_{i}, M)}{|V|},
\end{equation}
where $|V|$ denotes the number of frames in $V$, $V_{i}$ denotes the $i^{th}$ frame, and $d(V_{i}, M)$ is defined as:
\begin{equation}
d(V_{i}, M) = 1 - \frac{\sum_{x=1}^{W} \sum_{y=1}^{H} |V_{i}[x,y] - m[x,y]|}{W \cdot H},
\end{equation}
where $W$ and $H$ denote the width and height of the frame and the pixel values of the computed detection and the mask $m$ are 1 for water and 0 otherwise.

The classification-by-selection task is a more lenient task; given a selected area in a video, determine whether that area is a water surface or not. Although not as informative as the detection task, this task does offer several insights; it serves its own set of applications, such as human-aided water detection (i.e.\ water detection where the user specifies an interesting region). Also, it opens up the possibility for comparison against works from fields such as material and dynamic texture recognition.

\subsection{The Video Water Database} \label{sec:vwd}
Due to the lack of attention given to the specific task of water detection in videos, no database is available with a large enough quantity and variety for desirable evaluation. Therefore, the Video Water Database (VWD) is introduced here. This database contains several hours of video material of a wide range of water and non-water scenes. To the best of our knowledge, this is the largest database with video material on water.

%%% VERSION BEFORE REVISION
%In total, the database consists of 260 videos, where each video contains between 750 and 1500 frames, all with a frame size of $800 \times 600$. The water class consists of 160 videos of pre-dominantly 7 subcategories; canals, fountains, lakes, oceans, ponds, rivers, and streams. The non-water class can be represented by any other scene. Here, the non-water class is dominated by subcategories with seemingly similar spatial and temporal characteristics; clouds/steam, fire, flags, trees, and vegetation. An example of each of the subcategories in the database is shown in Fig.~\ref{fig:vwd}. All the videos are taken with a static camera, i.e.\ there are no large camera motions. In order to compute the quality of the computed detections, a binary mask is created for each video stating which pixels are water and which pixels are not. Care has furthermore been taken to maintain a large variety in scale and orientation of the water and non-water surfaces.
In total, the database consists of 260 videos, where each video contains between 750 and 1500 frames, all with a frame size of $800 \times 600$. The water class consists of 160 videos of pre-dominantly 7 subcategories; canals, fountains, lakes, oceans, ponds, rivers, and streams. The non-water class can be represented by any other scene. Here, the non-water class contains subcategories with similar spatial and temporal characteristics; clouds/steam, fire, flags, trees, and vegetation. An example of each of the subcategories in the database is shown in Fig.~\ref{fig:vwd}. All the videos are taken with a static camera, i.e. there are no large camera motions. Static cameras are employed here to be able to investigate the temporal and spatial properties of water in isolation. It furthermore allows us to quantify the performance of our hybrid descriptor. In order to compute the quality of the computed detections, a binary mask is created for each video stating which pixels are water and which pixels are not. Care has furthermore been taken to maintain a large variety in scale and orientation of the water and non-water surfaces.

%%% VERSION BEFORE REVISION
%The use of a static camera makes it possible to investigate the quality of the descriptors. Note however that the algorithm does not explicitly assume that the camera is static. Given that the pre-processing step and the temporal descriptor use a specific temporal window, camera motion can cause a temporary disruption of the results. Once the camera motion is finished, the algorithm will then automatically recover.
We note that our algorithm does not explicitly assume a static camera. Both the pre-processing and the temporal descriptor assume a temporal window at a specific spatial location. The temporal window in turn forms a trade-off; a smaller temporal window increases the robustness to camera motion, at the cost of discriminative power. 

Besides evaluating on the Video Water Database, a subset of 75 videos from the DynTex database~\cite{pet10} is also used for evaluation. The motive for this evaluation is two-fold. First, it shows that the algorithm is not tailored to the created database. Second, it provides a comparison for water detection against other non-water textures and objects. The selected subset contains humans, animals, traffic, windmills, flowers, and cloths. Since most of the named textures and objects will not be seen during training, the effectiveness of the algorithm on the DynTex database will provide insight into the generalization properties to unseen negatives.

\begin{table}[t]
\centering
\footnotesize
\begin{tabular}{l r r r r}
\hline
 & \multicolumn{4}{c}{No regularization}\\
 & \emph{Temporal} & \emph{Spatial} & \emph{Late Fusion} & \emph{Early Fusion}\\
\hline
Water & 90.4 & 85.5 & 90.7 & \textbf{90.8}\\
Non-water & 67.3 & 86.7 & 85.9 & \textbf{92.1}\\
Average & 78.9 & 86.1 & 88.3 & \textbf{91.5}\\
\hline
\end{tabular}\\
\vspace{0.25cm}
\begin{tabular}{l r r r r}
\hline
 & \multicolumn{4}{c}{Spatio-temporal regularization}\\
 & \emph{Temporal} & \emph{Spatial} & \emph{Late Fusion} & \emph{Early Fusion}\\
\hline
Water & 92.0 & 87.1 & 91.4 & \textbf{92.3}\\
Non-water & 68.6 & 89.8 & 90.7 & \textbf{95.0}\\
Average & 80.3 & 88.4 & 91.1 & \textbf{93.7}\\
\hline
\end{tabular}
\caption{Overview of the detection results of the descriptors and their fusions, resp. without and with regularization.}
\label{tab:overview1}
\end{table}

\subsection{Implementation details}
For the temporal descriptor, the length of the signal constitutes a trade-off between discriminative prowess and practicality. As the focus here is on detection and accuracy, a signal length of $m=200$ is used, with a resulting 200D feature vector. When combining the temporal and spatial descriptors before classification, the 200D temporal descriptor and 256D spatial descriptor are simply concatenated.

During training, 10 samples are retrieved from random locations for each frame of each training video, yielding roughly 750.000 samples to be trained by the Decision Forest. The main parameters of the Forest - the randomness and the number of trees - are set through validation~\cite{gem09}. For a test video, samples are extracted every 11$^{th}$ pixel in width and height for each frame, followed by individual classification. For the regularization, an equal contribution of the unary and pairwise terms (i.e.\ $\lambda=1$) has empirically shown to be most effective.

To evaluate the effectiveness of the algorithm for water detection, the primary components -- temporal and spatial features extraction, fusion, regularization -- are evaluated on the newly introduced Video Water Database. For this, the database is split randomly with equal ratio into a train and test split; equally among all subcategories.

\subsection{Water detection in the Video Water Database}
In Table~\ref{tab:overview1}, an overview is provided of the detection results for the algorithm without and with regularization. Individually and without regularization, the temporal and spatial descriptors yield 78.9\% and 86.1\% detection accuracy. It is interesting to observe that the water descriptor yields good performance for water, while the spatial descriptor yields good performance for non-water. The complementary nature also comes back in the performance of the fusions of the descriptors. The best performance is achieved by performing early fusion, with an increase to 91.5\% average detection rate. Early fusion is preferred here because of the large difference in true and false rates of the individual descriptors. In late fusion, the mistakes of an individual descriptor greatly influences the final detection result (due to the equal weighting of the probabilities). Early fusion however makes it possible to compensate for each others mistakes during training.

\begin{figure*}[t]
\centering
\begin{subfigure}{\linewidth}
\centering
\includegraphics[width=0.15\textwidth]{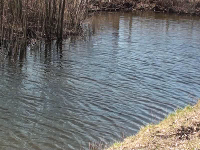}
\includegraphics[width=0.15\textwidth]{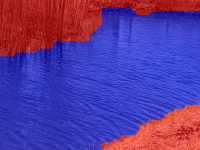}
\includegraphics[width=0.15\textwidth]{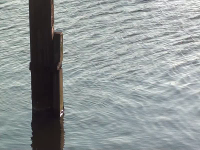}
\includegraphics[width=0.15\textwidth]{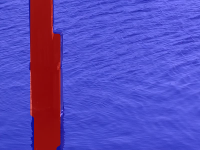}
\includegraphics[width=0.15\textwidth]{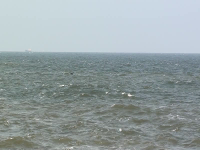}
\includegraphics[width=0.15\textwidth]{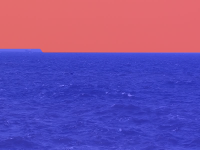}
\vspace{0.01cm}\\
\includegraphics[width=0.15\textwidth]{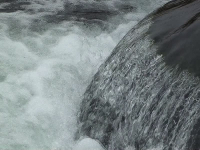}
\includegraphics[width=0.15\textwidth]{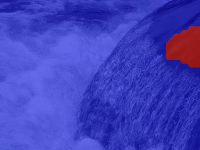}
\includegraphics[width=0.15\textwidth]{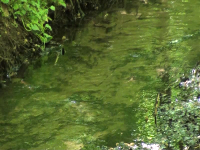}
\includegraphics[width=0.15\textwidth]{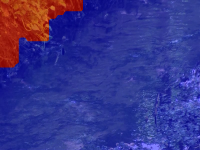}
\includegraphics[width=0.15\textwidth]{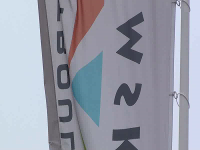}
\includegraphics[width=0.15\textwidth]{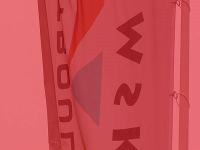}
\caption{Video Water Database.}
\label{fig:vwdres}
\end{subfigure}
\begin{subfigure}{\linewidth}
\centering
\includegraphics[width=0.15\textwidth]{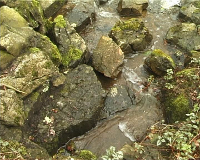}
\includegraphics[width=0.15\textwidth]{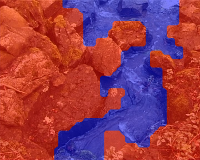}
\includegraphics[width=0.15\textwidth]{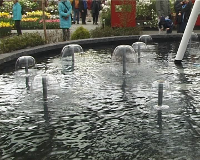}
\includegraphics[width=0.15\textwidth]{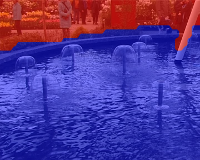}
\includegraphics[width=0.15\textwidth]{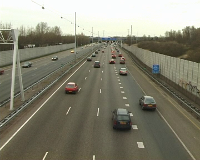}
\includegraphics[width=0.15\textwidth]{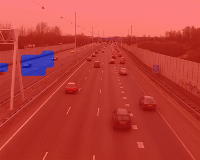}
\vspace{0.01cm}\\
\includegraphics[width=0.15\textwidth]{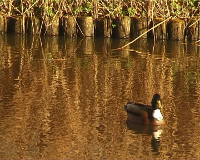}
\includegraphics[width=0.15\textwidth]{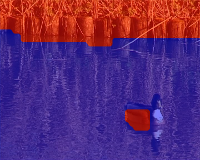}
\includegraphics[width=0.15\textwidth]{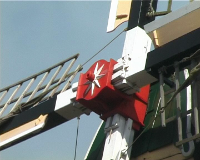}
\includegraphics[width=0.15\textwidth]{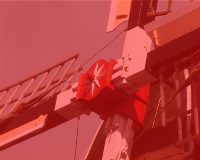}
\includegraphics[width=0.15\textwidth]{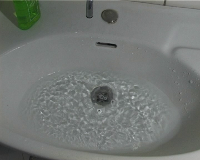}
\includegraphics[width=0.15\textwidth]{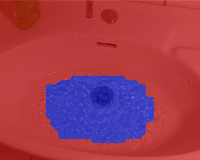}
\caption{DynTex database.}
\label{fig:dyntexres}
\end{subfigure}
\caption{Examples of detection results shown for both databases. Blue indicates water, red indicates non-water.}
\label{fig:visualres}
\end{figure*}

Next to fusing temporal and spatial information for local water classification, Table~\ref{tab:overview1} also indicates the effectiveness of regularization. Enforcing label consistency among spatio-temporal cliques removes classifier outliers and results in a smooth final detection result. The combination of the hybrid descriptor and spatio-temporal regularization yields a final detection accuracy of 93.7\%. In Fig.~\ref{fig:visualres}, the final detection result is shown for a number of test videos.

\begin{figure}[t]
\centering
\begin{subfigure}{\linewidth}
\centering
\includegraphics[width=0.24\textwidth]{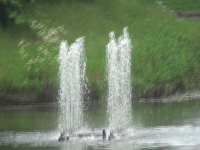}
\includegraphics[width=0.24\textwidth]{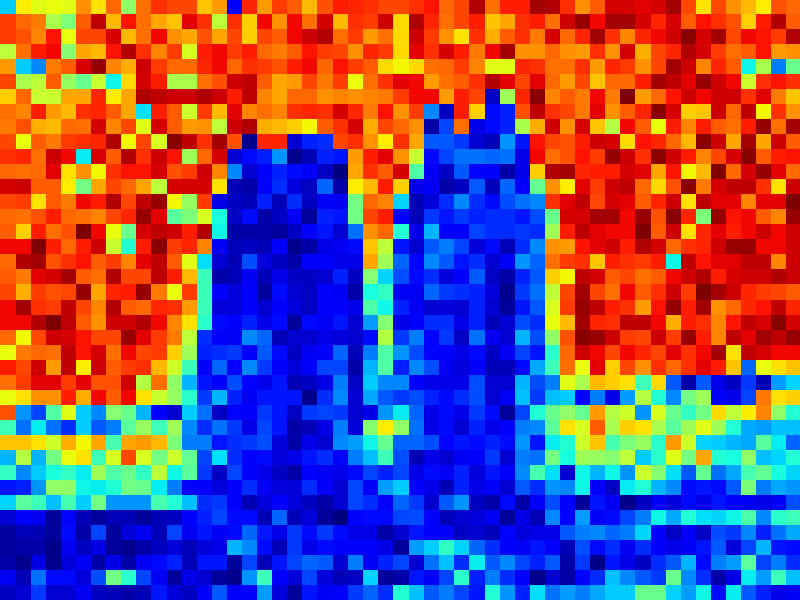}
\includegraphics[width=0.24\textwidth]{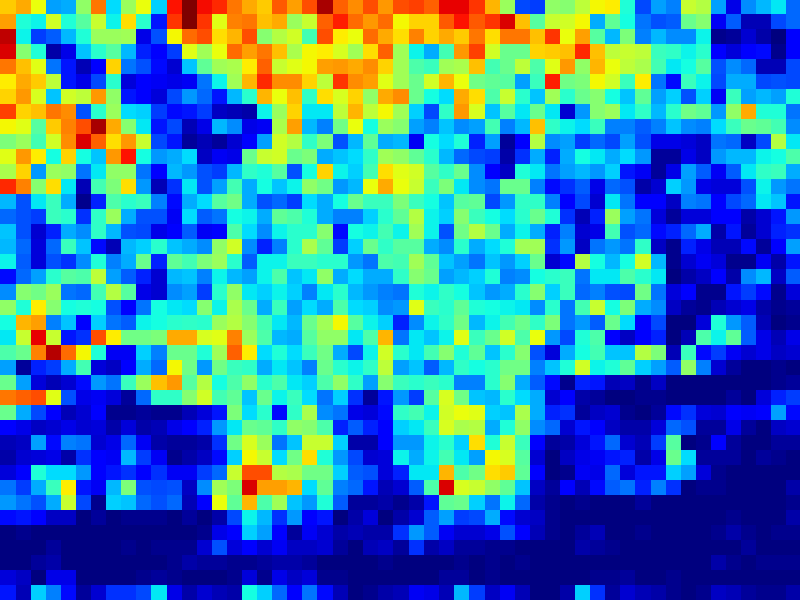}
\includegraphics[width=0.24\textwidth]{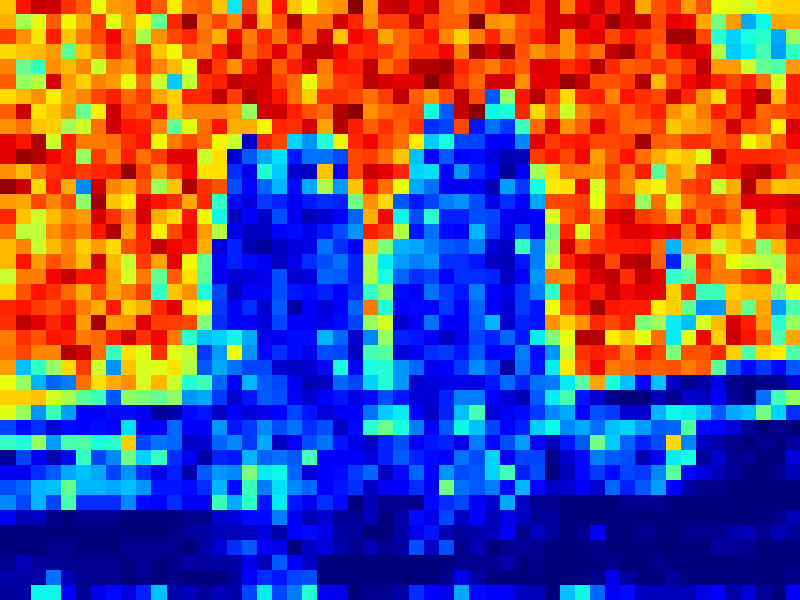}
\caption{Fountain.}
\label{fig:vwdcompa}
\end{subfigure}
\begin{subfigure}{\linewidth}
\centering
\includegraphics[width=0.24\textwidth]{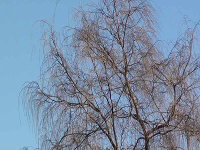}
\includegraphics[width=0.24\textwidth]{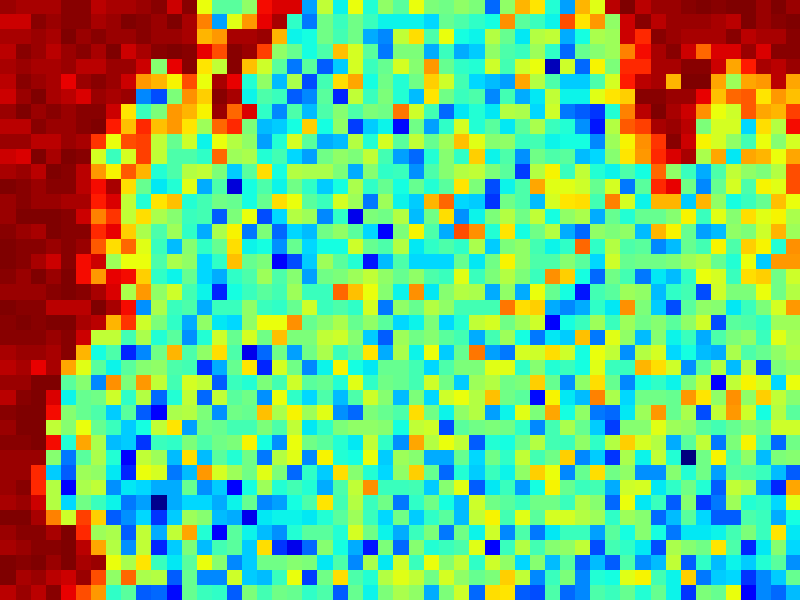}
\includegraphics[width=0.24\textwidth]{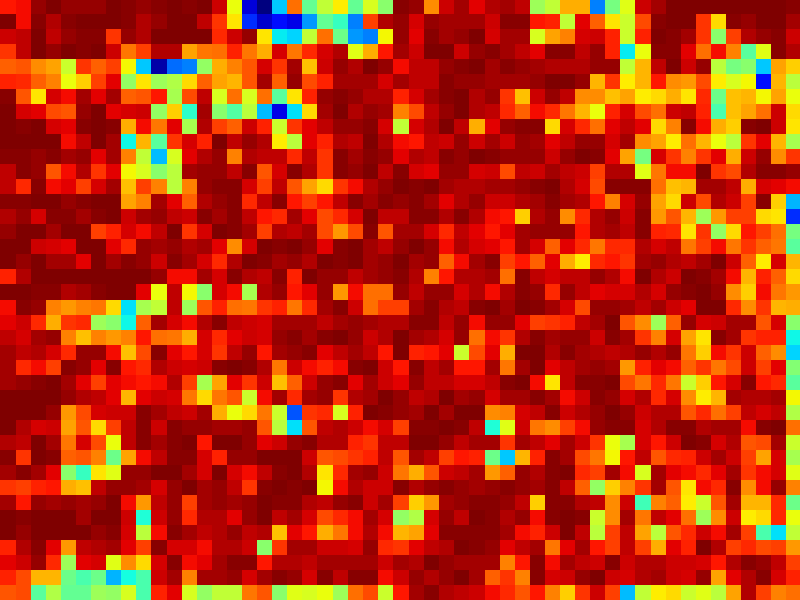}
\includegraphics[width=0.24\textwidth]{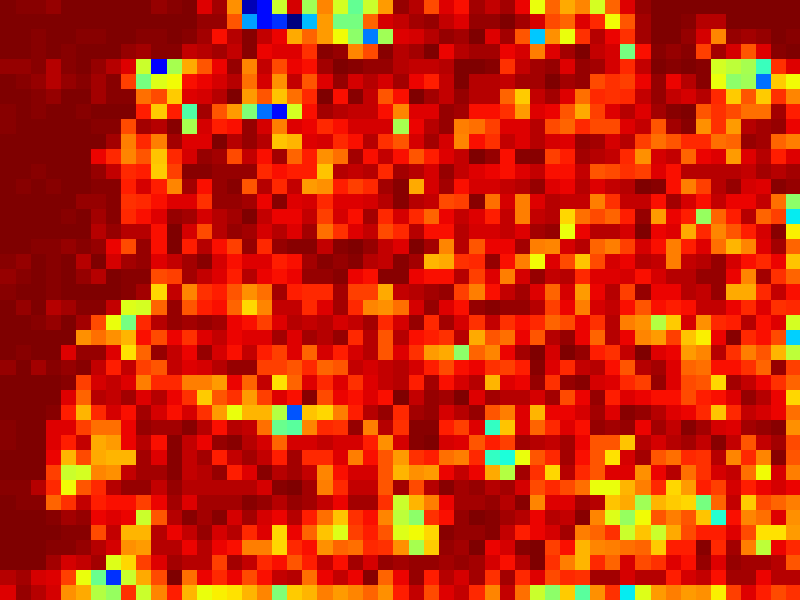}
\caption{Tree.}
\label{fig:vwdcompb}
\end{subfigure}
\caption{Two frames indicating the complementary nature of the temporal and spatial information. The first column shows a frame of the video. The second column shows the probability map of the temporal descriptor, the third column shows the map for the spatial descriptor, and the fourth column shows the map for the hybrid descriptor.}
\label{fig:vwdcomp}
\end{figure}

Interestingly, the strong increase in performance for the hybrid descriptor is not because of a strong increase in true detection rate. Contrarily, it is the false detection rate that achieves a strong decrease; from 32.7\% (temporal) and 13.3\% (spatial) to 5\%. This indicates that non-water elements might resemble water temporally or spatially, but not always spatio-temporally. The reasoning behind this observation is for a substantial part captured in Fig.~\ref{fig:vwdcomp}. In this Figure, two frames of test videos are shown, as well as the probability maps. In the probability maps, a blue colour indicates water, while a red colour indicates non-water. In Fig.~\ref{fig:vwdcompa}, it is shown that the temporal descriptor can aid the spatial descriptor, while Fig.~\ref{fig:vwdcompb} shows that the spatial descriptor can aid the temporal descriptor.

\subsection{Water detection in the DynTex Database}
To further emphasize the effectiveness of the introduced algorithm and in order to investigate the generalization properties of the algorithm, the water detection is also performed on a subset of the DynTex database~\cite{pet10}. In total, 75 videos of water and non-water scenes are selected. For training the model, the train split of the Video Water Database is used, while the model is evaluated on all the selected videos from the DynTex database.

Since the videos form the DynTex database have been captured with a different intent than the Video Water Database, different water and non-water types are present in this subset. For water, elements such as drinking water and water surfaces during rainfall are present. For non-water, new elements include windmills, animals, humans, and traffic. As these elements are not present in the training videos of the Video Water Database, a proper detection and classification of these videos greatly depends on the generalizing properties of the algorithm.

In Table~\ref{tab:overview2}, an overview is provided of the detection accuracies yielded on the DynTex subset. Although the numbers of Tables~\ref{tab:overview1} and~\ref{tab:overview2} are not directly comparable, the comparison does provide an indication of the generalization properties of the algorithm. Individually, the descriptors yield a lower detection accuracy on the DynTex database subset. However, the early fusion into the hybrid water descriptor results in a substantial boost from 76.7\% (temporal) and 77.9\% (spatial) to 91.3\% detection accuracy on average.

\begin{table}[h]
\centering
\begin{tabular}{l r r r r}
\hline
 & \emph{Temporal} & \emph{Spatial} & \emph{Hybrid}\\
\hline
Water & 89.7 (87.5) & 70.0 (68.4) & \textbf{87.9} (83.0)\\
Non-water & 63.7 (64.3) & 85.7 (79.6) & \textbf{94.7} (89.5)\\
Average & 76.7 (75.9) & 77.9 (74.0) & \textbf{91.3} (86.2)\\
\hline
\end{tabular}
\caption{Overview of the detection results on the DynTex database subset. The numbers within parentheses represent the detection results without regularization.}
\label{tab:overview2}
\end{table}

Fig.~\ref{fig:dyntexres} shows exemplary detections from the DynTex subset. For multiple examples, no similar video is present in the training set, e.g.\ the drinking water in the sink, the windmill, and the traffic. This Figure paints a similar picture to the results of Table~\ref{tab:overview2}; the algorithm is able to generalize to previously unseen water and non-water subcategories. This result highlights the goal of the algorithm to capture the inherent properties of water.

\subsection{Water classification-by-selection}
As a proof of concept and in order to compare the algorithm to a number of related papers, binary water classification is also considered. Here, the goal is to determine whether a video supplemented with a binary mask is water or not. The same training and testing splits are used as the detection task, while the manually created binary masks serve as binary masks to determine the foreground region.

For the introduced algorithm, the classification of a video is a function of the ratio of water and non-water pixels in the foreground region. Agnostic to any prior on the ratio of water and non-water pixels, a video is classified as water if the ratio of water pixels is at least~$\frac{1}{2}$, otherwise it is classified as non-water.

The classification accuracy of the algorithm is compared to multiple generic baselines from material and dynamic texture classification. In total, 4 algorithms are used as baseline methods. These baselines serve as general indicators of the complexity of the problem. The baseline algorithms include Volume Local Binary Patterns~\cite{zha07}, Linear Dynamical Systems~\cite{dor03}, Gabor filter bank distributions~\cite{var05}, and optical flow statistics~\cite{faz07}. For Volume LBP~\cite{zha07}, a $2^{14}$-dimensional feature vector is generated for a video by means of histogram binning using the 14 direct temporal and spatial neighbours of sampled foreground pixels. For LDS, the whole video has to be used, as the number of pixels needs to match between a pair of videos. Here, the setup of Saisan et al.~\cite{sai01} is followed.

For the filter bank distribution~\cite{var05}, a representation is generated by convolving frames with the rotation invariant MR8 filter bank~\cite{var05} and performing Vector Quantization (Bag-of-Words) on sampled response vectors. For the last baseline, 4 flow statistics are computed on estimated flows and averaged over the video. These statistics include characteristic direction, characteristic magnitude, divergence, and curl~\cite{faz07}. The optical flow baseline is performed both using the flow algorithm of Lucas and Kanade~\cite{luc81} and using the flow algorithm of Horn and Schunck~\cite{hor81}.

An overview of the classification accuracies is highlighted in Table~\ref{tab:overview3}. On the Video Water Database, the hybrid descriptor outperforms both the individual descriptors (similar to water detection) and the baseline methods. In fact, the only baseline method that comes near the results of the hybrid descriptor is Volume LBP~\cite{zha07}. This result is not entirely surprising, as the purely spatial variant is pursued throughout this work as the spatial descriptor. Because of the descriptor length of Volume LBP (either $2^{14}$ or $2^{26}$), it is impractical to use it as a local descriptor in the proposed water algorithm. All other baselines do not meet the performance of the hybrid descriptor.

\begin{table}[t]
\centering
\begin{tabular}{l r r r}
\hline
\emph{Methods} & \emph{VWD} & \emph{Dyntex} & \emph{Abs. diff.}\\
\hline
\emph{Ours, hybrid} & \textbf{98.4} & \textbf{95.8} & \textbf{-2.6}\\
\emph{Ours, spatial} & 93.1 & 84.6 & -8.5\\
\emph{Ours, temporal} & 83.0 & 81.0 & -2.0\\
\hline
Volume LBP~\cite{zha07} & 93.8 & 79.1 & -14.7\\
MR8 filter bank~\cite{var05} & 84.3 & 67.2 & -17.1\\
Flow stats (HS)~\cite{faz07,hor81} & 75.0 & 55.4 & -19.6\\
LDS~\cite{dor03} & 67.4 & 56.3 & -11.1\\
Flow stats (LK)~\cite{faz07,luc81} & 62.8 & 49.7 & -13.1\\
\hline
\end{tabular}
\caption{Classification accuracy results yielded for both the Video Water Database (second column) and the Dyntex database (third column). The fourth column states the absolute difference in achieved accuracy between the Video Water Database and the Dyntex database.}
\label{tab:overview3}
\end{table}

As indicated in the third column of Table~\ref{tab:overview3}, all methods yield a lower classification accuracy on the Dyntex database. Although the numbers can not directly be compared to the numbers of the Video Water Database, the decline in performance of each of the methods provides a clear indication of the performance of the water algorithm. For the water algorithm, the hybrid and temporal descriptor indicate the best generalization capabilities, while the spatial descriptor reports a 8.5\% decline (absolute difference). For the baseline methods, a decline of 11.1\% and higher is reported. This result indicates that the introduced water algorithm not only outperforms the baseline methods, it is also able to generalize better to unseen water and non-water subcategories.

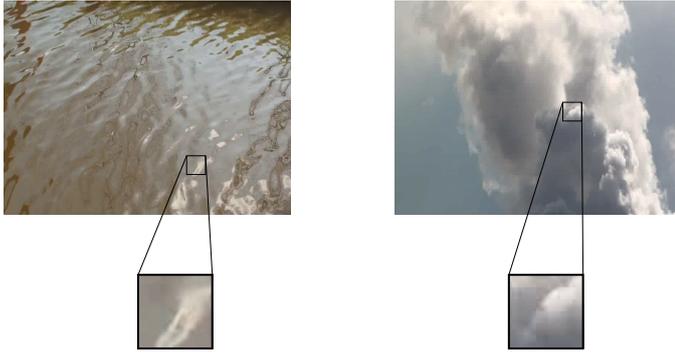
\begin{figure}[t]
\centering
\resizebox{0.49\textwidth}{!}{
\begin{tikzpicture}
\node[anchor=south west,inner sep=0] at (0,2) {\includegraphics[width=0.2315\textwidth]{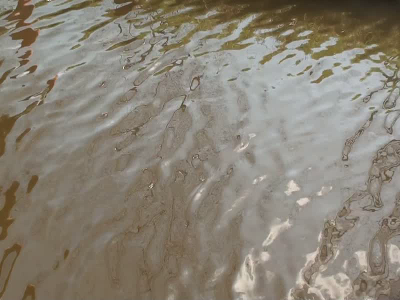}};
\draw [draw=black, line width=0.15mm] (2.72,2.6) rectangle (3,2.88);
\node[anchor=south west,inner sep=0] at (5.8,2) {\includegraphics[width=0.2315\textwidth]{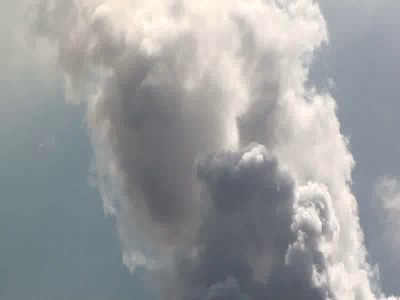}};
\draw [draw=black, line width=0.15mm] (8.3,3.4) rectangle (8.58,3.68);
\node[anchor=south west,inner sep=0] at (2,0) {\includegraphics[width=0.06\textwidth]{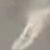}};
\draw [draw=black, line width=0.3mm] (2,0) rectangle (3.1,1.1);
\node[anchor=south west,inner sep=0] at (7.5,0) {\includegraphics[width=0.06\textwidth]{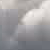}};
\draw [draw=black, line width=0.3mm] (7.5,0) rectangle (8.6,1.1);

\draw [draw=black, line width=0.15mm] (2.72,2.88) -- (2,1.1);
\draw [draw=black, line width=0.15mm] (3.00,2.88) -- (3.1,1.1);
\draw [draw=black, line width=0.15mm] (8.3,3.68) -- (7.5,1.1);
\draw [draw=black, line width=0.15mm] (8.58,3.68) -- (8.6,1.1);
\end{tikzpicture}}
\caption{Visual example indicating the complexity of water detection purely from local information.}
\label{fig:highlight}
\end{figure}

Both for the detection and classification tasks, it can be noted that the final numbers are rather high. This is first and foremost due to the nature of the task; it is cast as a strictly binary problem. This means that if a local video volume or even a whole video of a tree is classified as fire, there will be no loss. As long as the water/non-water boundary line is not crossed, no loss occurs. Note however that this hardly makes the problem easy, especially from a purely local perspective. When treating each local video volume independently for classification, any form of contextual information is discarded. This is highlighted in Figure~\ref{fig:highlight}. When looking at the whole frames, it is not hard to make out which one is water and which one is a cloud. However, purely based on the local squares, it becomes exceedingly harder to state which one is part of a water surface and which one is not. This indicates the complexity of a non-holistic approach to the detection problem.

\section{Conclusions}
In this work, the problem of detecting water in videos is tackled. As the specific problem of water detection has hardly been addressed in related work, this work investigates the temporal and spatial dynamics of water. First, a pre-processing stage is introduced that is aimed at removing reflections and water colours. After that, a hybrid descriptor and local detection algorithm are introduced for discovering water regions in a video. To evaluate the algorithm, the Video Water Database is furthermore introduced. Quantitative and qualitative evaluation show that the algorithm is able to robustly detect region of water in videos, with a high detection accuracy and a classification accuracy that outperforms algorithms from directly related fields.

\section*{Acknowledgement}
This work is supported by the FES project COMMIT.



%% References


\section*{References}
\bibliographystyle{elsarticle-num}
\bibliography{egbib.bib}






% \end{thebibliography}

\end{document}